\documentclass[letterpaper]{article} 
\usepackage{aaai2026}  
\usepackage{times}  
\usepackage{helvet}  
\usepackage{courier}  
\usepackage[hyphens]{url}  
\usepackage{graphicx} 
\usepackage{natbib}  
\usepackage{caption} 
\usepackage{algorithm}
\usepackage{algorithmic}
\usepackage{multirow}
\usepackage{newfloat}
\usepackage{listings}
\usepackage{amsmath}
\usepackage{amssymb}
\usepackage{soul}
\usepackage{bm}
\DeclareCaptionStyle{ruled}{labelfont=normalfont,labelsep=colon,strut=off} 
\floatstyle{ruled}
\newfloat{listing}{tb}{lst}{}
\floatname{listing}{Listing}
\usepackage{pdfpages}

\title{CodeFormer++: Blind Face Restoration Using Deformable Registration and Deep Metric Learning}

\author{
    Venkata Bharath Reddy Reddem\textsuperscript{\rm 1},
    Akshay P Sarashetti\textsuperscript{\rm 2},
    Ranjth Merugu\textsuperscript{\rm 3},
    Amit Satish Unde\textsuperscript{\rm 2}
}
\affiliations{
\textsuperscript{\rm 1}University of California, Irvine\\
\textsuperscript{\rm 2}Samsung R\&D Institute India, Bangalore\\
\textsuperscript{\rm 3}Stony Brook University\\

}

\usepackage{bibentry}

\begin{document}

\maketitle

\begin{abstract}
	Blind face restoration (BFR) has attracted increasing attention with the rise of generative methods.  Most existing approaches integrate generative priors into the restoration process, aiming to jointly address facial detail generation and identity preservation. However, these methods often suffer from a trade-off between visual quality and identity fidelity, leading to either identity distortion or suboptimal degradation removal. 
	In this paper, we present CodeFormer++, a novel framework that maximizes the utility of generative priors for high-quality face restoration while preserving identity.
	We decompose BFR into three sub-tasks: (i) identity-preserving face restoration, (ii) high-quality face generation, and (iii) dynamic fusion of identity features with realistic texture details. Our method makes three key contributions: (1) 
	a learning-based deformable face registration module that semantically aligns generated and restored faces; (2) a texture guided restoration network to dynamically extract and transfer the texture of generated face to boost the quality of identity-preserving restored face; and (3) the integration of deep metric learning for BFR with the generation of informative positive and hard negative samples to better fuse identity-preserving and generative features. 
	Extensive experiments on real-world and synthetic datasets demonstrate that, the proposed CodeFormer++ achieves superior performance in terms of both visual fidelity and identity consistency.  
\end{abstract}


\section{Introduction}

Blind Face Restoration (BFR) is a well-established problem in computer vision. Its primary objective is to reconstruct a high-quality (HQ) face image from a low-quality (LQ) input while preserving the identity. In real-world scenarios, face images are often affected by complex combinations of degradations such as blur, noise, and compression artifacts. These diverse degradations pose significant challenges for effective restoration. 

\begin{figure}[t]
	\centering
	\includegraphics[width=0.9\linewidth]{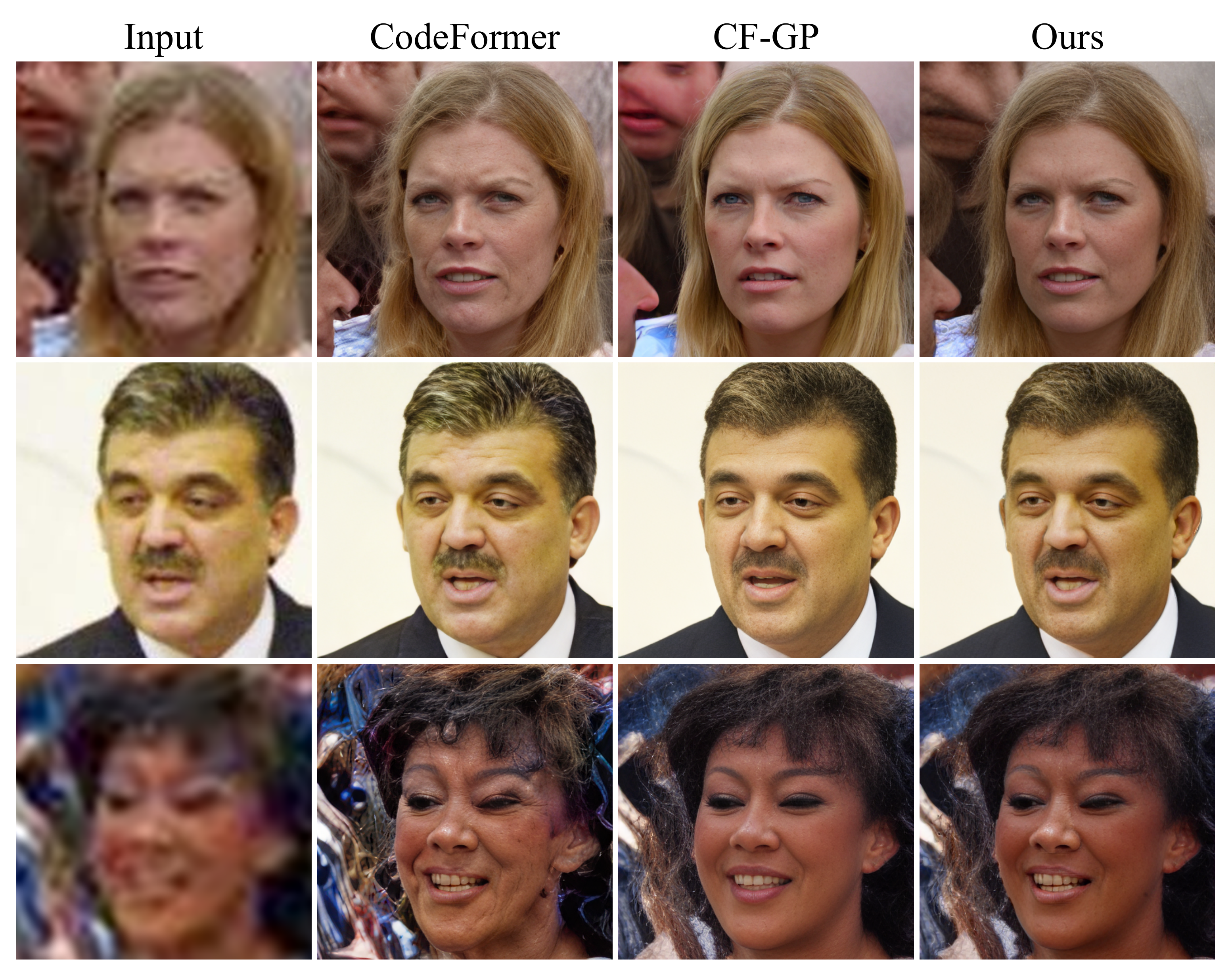}

	\caption{Given a degraded face image, our method is able to reconstruct a high-fidelity, texture-rich image. In contrast, CodeFormer fails to completely remove the degradation and tends to produce overly smoothed results. Although generative prior CF-GP generates images with realistic textures, it suffer from identity preservation issues.}
	\label{fig:frontFig}
\end{figure}

\begin{figure*}[t]
	\centering
	\includegraphics[width=0.9\textwidth]{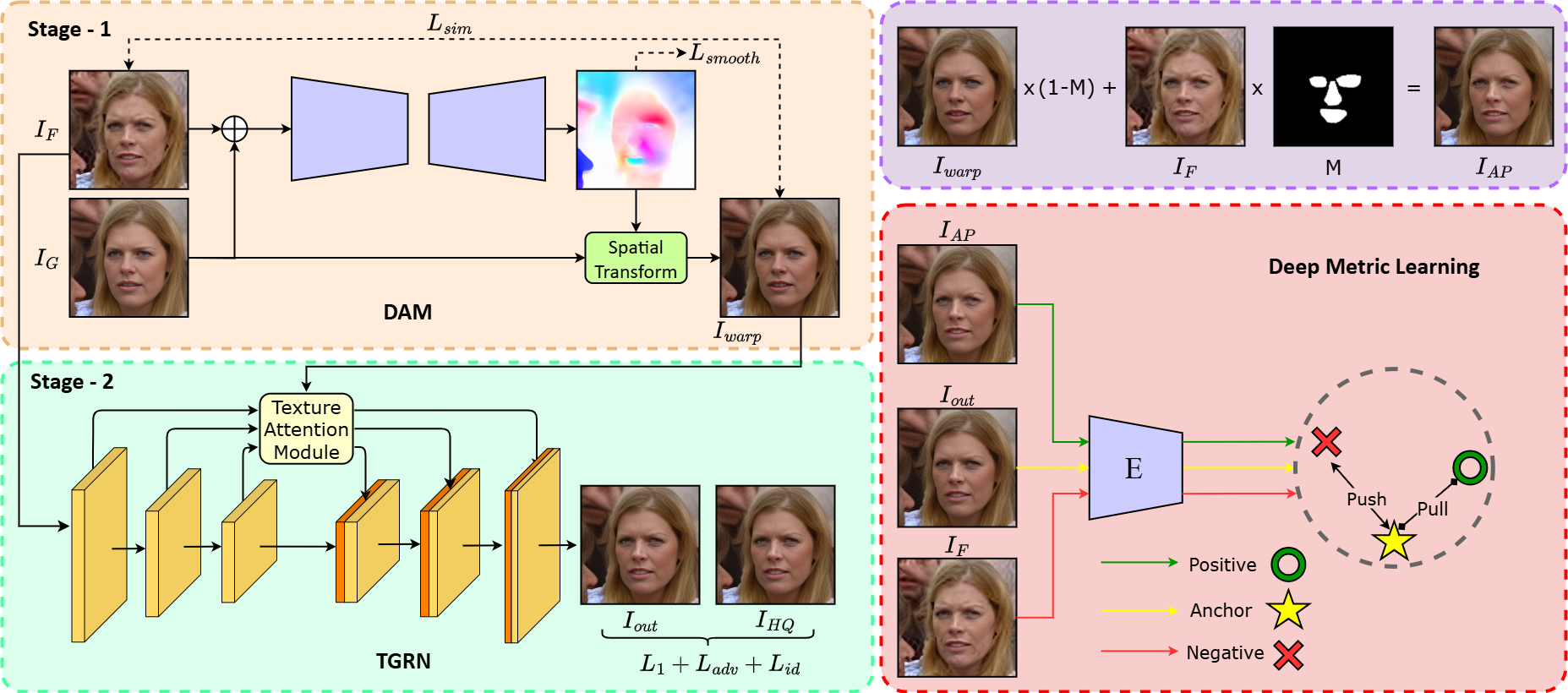}
	\caption{\textbf{Overview of our CodeFormer++ framework.} In stage-1, Deformable image Alignment Module (DAM) is trained to predict deformation field between $I_F$ and $I_G$. In stage-2, Texture-prior Guided Restoration Network (TGRN) is trained to
		generate texture-rich and high-fidelity output by injecting texture from $I_{warp}$. The hard positive sample $I_{AP}$ is obtained by combining facial components from
		$I_F$ and texture from $I_{warp}$ to enforce optimal balance between realism and fidelity.
		TGRN is supervised using deep metric learning to focus on extracting texture from $I_{AP}$ by pulling anchor towards positive image and away from negative image.}
	\label{fig:network}
\end{figure*}

In recent years, BFR is gaining attention of research community owing to significant advancement in powerful generative models. Recent approaches~\cite{wang2021towards, yang2021gan} exploit the powerful priors of pretrained face image generators such as StyleGAN~\cite{karras2019style} to improve robustness against real-world, unknown degradations. Albeit promising, these continuous latent space based methods suffer from poor fidelity. This is due to difficulty in finding the accurate latent vectors in infinite search space. To alleviate the challenges associated with continuous latent spaces, recent works~\cite{gu2022vqfr, zhou2022towards, tsai2023dual} leverage vector-quanitized codebook prior that encodes face images to a discrete latent space. The vector quantization mechanism reduces uncertainty in LQ-HQ mapping due to constrained search space, enhancing the robustness of these methods to various degradations. 

The codebook-prior based BFR approaches typically rely on two sources of information: the degraded input, which contains critical identity-preserving features, and a pretrained decoder as a prior, for the generation of high-quality face images. These existing methods attempt to jointly solve the challenges of identity preservation and texture generation within a single unified pipeline ~\cite{tsai2023dual, yue2024difface}. However, such approaches often struggle to balance these conflicting objectives. Methods that emphasize on high-quality synthesis~\cite{wang2021towards} often fail to preserve identity, while identity-focused approaches~\cite{zhou2022towards} typically yield over-smoothed results with limited texture diversity and inadequate degradation removal.

A similar trade-off between fidelity and quality is notably observed in CodeFormer~\cite{zhou2022towards}. The Controllable Feature Transformation (CFT) module is introduced to adjust the information flow from the LQ input to the restored output via a scalar weight $w \in [0,1]$. By varying $w$, the model can interpolate between identity fidelity and visual quality. Empirical observations reveal that increasing the dependency on the LQ image ($w=1$) improves identity preservation but at the cost of reduction in visual quality as shown in Fig. \ref{fig:frontFig}. This degradation is primarily due to the corrupted feature flow from the encoder, which becomes increasingly unreliable when the input suffers from complex artifacts. Conversely, reducing the scalar weight ($w\approx0$) minimizes reliance on the degraded input and leads to visually appealing results. However, this often comes at the expense of inconsistent identity, as the generated outputs exhibit noticeable semantic shifts in key facial regions such as the jawline, eyes, nose, and mouth (see Fig. \ref{fig:frontFig}). These observations highlight the inherent difficulty in jointly optimizing for fidelity and perceptual quality within a unified framework. The existing methods struggle to simultaneously achieve both constraints, motivating the need for a more principled and modular approach to face restoration.

In this paper, we propose CodeFormer++, a novel face restoration framework that dynamically fuses identity-preserving low-quality facial features, with high-quality but identity-altered generative features. We aim to address the critical challenge observed in CodeFormer, where the balance between identity preserving and generative features remains suboptimal. Unlike conventional methods, we decompose the problem into four key stages: 1) Identity-preserving face restoration (CFT with $w=1$), referred as \textbf{CF-ID}; 2) High-quality face image generation as a prior (CFT with $w=0$), referred to as \textbf{CF-GP}; 3) Deformable alignment of CF-GP towards CF-ID image using an optical flow to reduce structural bias between them; and 4) Dynamic fusion of identity information with realistic texture details through incorporation of deep metric learning into our pipeline.

Our main contributions are summarized as follows:
\begin{itemize}
	\item We propose a novel and generic framework for synergistically fusing identity-preserving features and generative priors, enabling high-fidelity face restoration with rich perceptual detail.
	
	\item We present Deformable image Alignment Module (DAM) for semantically aligning CF-ID and CF-GP images by establishing dense, non-linear correspondence between them. 
	
	\item We introduce a Texture-prior Guided Restoration Network (TGRN) with deep metric learning to ensure that the restored face inherits texture from CF-GP image, while remaining semantically aligned with CF-ID image.
	
	\item We also propose a novel hard sampling strategy for deep metric learning to enforce optimal balance between realism and fidelity. 
	
	\item Extensive experimental studies to demonstrate that our proposed method outperforms state-of-the-art (SOTA) approaches on both synthetic and real-world datasets, exhibiting superior performance in terms of perceptual quality and identity preservation.
\end{itemize}

%
%
\section{Related Work}
Recent methods explored generative priors for BFR. Generative adversarial network (GAN)  based solutions like PULSE~\cite{menon2020pulse}, mGANprior~\cite{gu2020image}, GFPGAN~\cite{wang2021towards} and GPEN~\cite{yang2021gan} primarily use StyleGAN~\cite{karras2019style} as their generative backbone, owing to their remarkable capacity in synthesizing high-quality facial images. Since the emergence of diffusion models as powerful tools for generating high-quality realistic images, diffusion based methods~\cite{yue2024difface, wang2023dr2} have also been explored for BFR. These approaches operate in continuous latent space and  heavily rely on latent codes estimation from the LQ image. Despite generating visually plausible faces, they often lack in accurately restoring identity features. 

To improve upon this, latest methods explore codebook based priors~\cite{van2017neural,esser2021taming}, where a codebook of quantized embeddings is learnt to represent high-quality features. These learnings are constrained to discrete latent space. Vector Quantized Variational Autoencoders (VQ-VAE)~\cite{van2017neural} introduced a discrete latent space using a learned codebook of quantized embeddings. Following these developments, several works~\cite{gu2022vqfr, wang2022restoreformer} have utilized vector-quantized codebooks to generate high-quality facial images. While these methods achieve impressive perceptual realism, they often struggle to maintain identity consistency with the input images. DAEFR~\cite{tsai2023dual} proposed learning separate codebook priors for LQ and HQ images to reduce domain gap. Although it achieves improved perceptual quality, it suffers from inadequate spatial-level conditioning from the encoder, resulting in subpar identity preservation.

These limitations underscore the need for a framework that can dynamically fuse semantically aligned generative features and identity-preserving cues, which we address in our proposed CodeFormer++ architecture.

\section{Methodology}
\subsection{Overall Framework}
The primary goal of this work is to synergize generative priors, CF-GP with identity-preserving features, CF-ID to achieve high-quality, high-fidelity face restoration. An overview of the proposed architecture is shown in Fig. \ref{fig:network}. We first introduce the deformable alignment module, which semantically aligns CF-GP to CF-ID, thereby reducing structural mismatch between them and facilitating coherent fusion. Following alignment, we propose a texture-prior guided restoration network to enrich CF-ID with fine-grained textures from the aligned CF-GP image. TGRN integrates a Texture Attention Module (TAM) to dynamically fuse texture-rich features from aligned CF-GP with identity cues from CF-ID.  Furthermore, to reinforce identity preservation and guide the network toward a more discriminative feature space, we incorporate a deep metric learning objective using a triplet-based contrastive loss. This supervision encourages the model to restore face images that are perceptually realistic and identity-consistent. The following sections describe each component of CodeFormer++ in more detail.

\begin{figure}[t]
	\centering
	\includegraphics[width=0.85\linewidth]{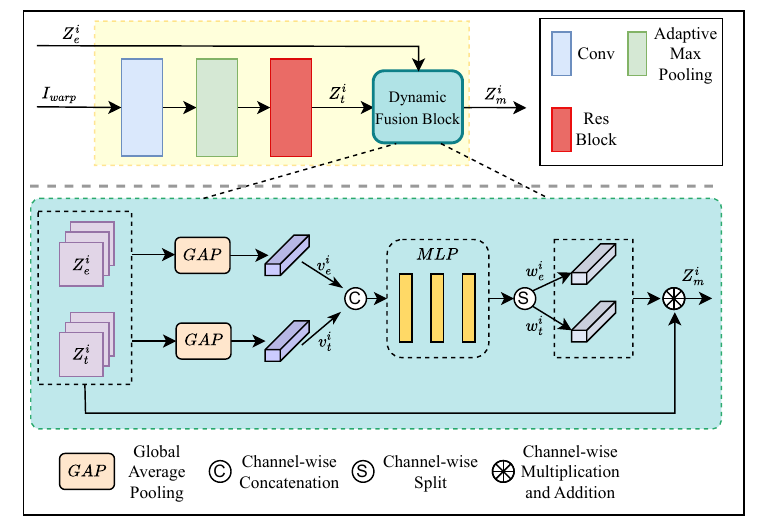}
	\caption{The architecture of texture attention module.}
	\label{fig:tam}
\end{figure}

\subsection{Deformable Image Alignment Module}
We refer high-fidelity face image CF-ID and high-quality but identity altered face image CF-GP as  $I_F$ and $I_G$, respectively. Due to their disparate objectives, these two outputs often exhibit severe semantic misalignment, particularly in facial structures such as the jawline, nose, and mouth. This structural discrepancy hinders direct fusion of their respective strengths, texture from $I_{G}$ and fidelity from $I_{F}$.

To mitigate this, we introduce the deformable image alignment module by formulating the alignment problem as a deformable image registration task. Similar to joint refinement strategies employed in video restoration frameworks such as ~\cite{merugu2025joint}, which iteratively refine optical flow and feature representations for temporal coherence, our approach couples alignment and feature consistency through a learnable deformation field. Inspired by the VoxelMorph~\cite{balakrishnan2019voxelmorph} framework (originally developed for 3D medical image registration), we model 2D face image alignment through a learnable function $R_{\theta}(I_F, I_G)$ that predicts a dense deformation field $\phi$, aligning $I_{G}$ with $I_{F}$. It is mathematically represented as:

\begin{equation}
	R_{\theta}(I_F, I_G) = \phi
\end{equation}
where $\theta$ denotes learnable network parameters. 

Using the deformation field $\phi$, we then warp $I_{G}$ towards $I_{F}$ using a differentiable spatial transformation layer, resulting in an aligned image $I_{warp}$. This warped image retains the rich texture and perceptual quality of  $I_{G}$, while being structurally consistent with the identity-preserving $I_{F}$. DAM facilitates a more coherent and effective fusion of texture and identity information in subsequent stages, enabling improved restoration of both visual realism and semantic consistency.

\subsection{Texture-Prior Injection in Restoration Network}
Once the generative prior is semantically aligned with the identity-preserving image $I_{F}$, our objective is to effectively inject the rich facial texture from $I_{warp}$ into $I_{F}$. To achieve this, we propose the texture-prior guided restoration network, which is a specialized architecture designed to harmonize fidelity with perceptual quality. TGRN comprises two key components: 1) a U-Net backbone for structural restoration, and 2) a texture attention module for adaptive fusion of identity and texture cues.

We adopt a three-level U-Net encoder-decoder architecture, using $I_{F}$ as the primary input to preserve identity-specific features. The encoder extracts multi-scale feature representations from $I_{F}$, where features at the i-th encoder level are denoted as $Z^{i}_{e} \in  \mathbb{R}^{m \times n \times d}$, with $m$, $n$, and $d$ representing spatial height, width, and channel dimensions, respectively.

Parallel to this, the aligned prior $I_{warp}$ is processed through the TAM. It consists of convolutional layers, adaptive max pooling, and residual blocks to extract texture-aware features $Z^{i}_{t}$ at each semantic level $i$. The adaptive max pooling layer ensures that the spatial resolution of $Z^{i}_{t}$ matches that of the corresponding $Z^{i}_{e}$, enabling effective feature alignment and fusion. To inject texture information selectively while maintaining structural fidelity, we utilize a dynamic fusion block at each encoder level. This block first applies global average pooling to both feature maps $Z^{i}_{e}$ and $Z^{i}_{t}$ to obtain compact global descriptors:
\begin{equation}
	\begin{split}
		v^i_e =\frac{1}{m \times n} \sum_{s=1}^{m} \sum_{t=1}^{n}Z^i_e(s, t) \\
		v^i_t =\frac{1}{m \times n} \sum_{s=1}^{m} \sum_{t=1}^{n}Z^i_t(s, t)  \label{eq:global_dynamic}
	\end{split}
\end{equation}
These global features are concatenated and passed through a Multi-Layer Perceptron (MLP) to estimate channel-wise dynamic fusion weights:
\begin{equation}
	\left[w^{i}_{e}, w^{i}_{t}\right]=\text{MLP}(\left[v^i_{e}, v^i_{t}\right])
\end{equation}
where MLP consists of three fully connected layers and outputs $w^{i}_{e}, w^{i}_{t} \in \mathbb{R}^{{d}}$ represent the learned weights for identity and texture features respectively. Using these weights, the final fused representation $Z^i_m$ is computed as:
\begin{equation} 
	Z^i_m =  w^{i}_{e}\odot Z^i_e+ w^{i}_{t}\odot Z^i_t
	\label{eq:global_dynamic_final}
\end{equation}
where $\odot$ denotes channel-wise multiplication. These fused features $Z^i_m$ are passed through the decoder to reconstruct the output image, effectively enriching the facial details while preserving identity characteristics.

\subsubsection{Feature Fusion Objective via Deep Metric Learning}
The core objective of the TGRN is twofold: 1) retain identity-specific features from the high-fidelity $I_{F}$ and 2) enhance perceptual realism by transferring texture details from the generative prior $I_{warp}$. To ensure that the restored image maintains identity while benefiting from texture priors, we adopt a deep metric learning framework that explicitly guides the network using a contrastive embedding space. Traditional deep metric learning approaches rely on minimizing the distance between similar samples (positive pairs) and maximizing it for dissimilar samples (negative pairs). Existing methods typically use the ground-truth (GT) image as the positive sample. However, since our task is focused on fusing the information between $I_F$ and $I_{warp}$, GT is not a suitable candidate as a positive sample. To better guide identity-texture fusion, we propose a novel anchor-positive sample construction strategy. Specifically, we synthesize an anchor-positive image $I_{AP}$, by combining facial components (eyes, nose, and mouth) from the identity-preserving image $I_{F}$ and transferring the skin regions and contextual textures from $I_{warp}$, as illustrated in Fig. \ref{fig:network} and Eq. \ref{eq:combination}.

\begin{equation}
	I_{AP} = I_{F}*M + I_{warp}*(1-M),
	\label{eq:combination}
\end{equation}
where $M$ is a binary semantic map~\cite{yasarla2020deblurring} highlighting facial regions crucial for identity.

Conventional settings for deep metric learning often designate the LQ image as the negative sample. However, such positive-negative pairs are trivially separable, weakening the discriminative power of the learned embedding. Instead, inspired by hard negative mining strategies~\cite{chuang2020debiased}, we select $I_F$ itself as a hard negative, since it shares identity structure but lacks the enhanced perceptual texture of the restored image. By using $I_F$ as negative sample and $I_{AP}$ as positive sample, we enforce the network to induce realistic texture on the high-fidelity output image. We employ a cosine triplet loss to supervise feature embedding distances, using triplets ($f_{p}, f_{a}, f_{n}$) extracted using a pretrained VGG network from $I_{AP}$, $I_{out}$, and $I_F$ respectively.
The cosine-based triplet loss is defined as:
\begin{equation}
	L_{triplet} = -log \frac{e^{f_{p}f_{a}}}{e^{f_{p}f_{a}} + e^{f_{n}f_{a}}},
	\label{eq:triplet}
\end{equation}
\begin{equation}
	f_{p}f_{a} = ||f_{p}||\hspace{0.1cm} ||f_{a}|| cos(\theta^+),
	\label{eq:dotproduct}
\end{equation}
The Eq. \ref{eq:dotproduct} represents the dot product between two vectors, wherein $\theta^+$ is the angle between the vectors.
Since all features are L2-normalized, $||f_{p}||=||f_{a}||=||f_{a}||=1$, the final formulation becomes:
\begin{equation}
	L_{triplet} = -\lambda_{triplet}\hspace{0.1cm} log \frac{e^{cos(\theta^+)}}{e^{cos(\theta^+)} + e^{cos(\theta^-)}},
	\label{eq:triplet2}
\end{equation}
We set triplet loss weight  $\lambda_{triplet}=1$ in our experiments.



\subsection{Training CodeFormer++}
We describe in the following training process and loss functions used for optimizing deformable image alignment module and texture-prior guided restoration network.

\noindent \textbf{Training DAM.} We train DAM using two losses: $L_{sim}$, which penalizes the difference in appearance, and $L_{smooth}$ that penalizes local spatial variations in $\phi$. We adopt the negative local normalized cross-correlation between the aligned image $I_G$($\phi$) and the high-fidelity image $I_F$, a widely used metric in registration tasks~\cite{meng2024correlation}. $\hat{I}_F(p)$ and $\hat{I}_G(\phi(p))$ denote the local mean intensities, where $p_i$ iterates over a local neighborhood $\Omega$ of size $n^2$ around point $p$, where n = 9 in our experiments. It is defined as:
\fontsize{7}{11}\selectfont
\begin{multline}
	L_{sim}(I_F, I_G(\phi)) = \\
	- \sum_{p\in \Omega} \frac{\left( \sum_{p_i} [I_F(p_i)-\hat{I}_F(p)][I_G(\phi(p_{i})) - \hat{I}_G(\phi(p))]\right)^{2}}{\left( \sum_{p_i}[I_F(p_i)-\hat{I}_F(p)]^{2} \right) \left(\sum_{p_i}[ I_G(\phi(p_{i}))-\hat{I}_G(\phi(p))
		]^{2} \right)},
\end{multline}
\fontsize{10}{11}\selectfont

To ensure spatial continuity in the deformation field $\phi$, we use:
\begin{equation}
	L_{smooth}(\phi) = \sum_{p\in \Omega} ||\nabla \phi(p)||^{2},
	\label{eq:smooth}
\end{equation}
where $\nabla$ is the spatial gradient operator. The overall loss for training deformable image alignment module is defined as:
\begin{equation}
	L(I_F, I_G, \phi) = L_{sim}(I_F,I_G(\phi)) + \lambda_{\phi} L_{smooth}(\phi),
	\label{eq:loss2}
\end{equation}
where $\lambda_{\phi}$ is the regularization parameter to balance the registration and transformation smoothness.

\noindent \textbf{Training TGRN.} 
To address cases where severe degradation leads to residual artifacts in CodeFormer outputs, we refine the restoration using TGRN, which is supervised using a combination of regression, adversarial, identity, and metric learning losses. The goal is to bring the output $I_{out}$ closer to the GT high-quality image $I_{HQ}$.

\begin{equation}
	\begin{split}
		L_1 = ||I_{HQ}-I_{out}||_1, \\
		L_{adv} = -\mathbb{E}_{I_{out}}\ \mathtt{softplus}(D(I_{out})), \\
		L_{id} =  \|\eta(I_{HQ})-\eta(I_{out}) \|_1,
	\end{split}
\end{equation}
where $D$ and $\eta$ denotes the discriminator and the ArcFace feature extractor respectively. 

The overall objective of the texture-prior guided restoration network is the combination of above losses:
\begin{equation}
	L_{total} = \lambda_{l1} L_1 + \lambda_{adv} L_{adv} + \lambda_{id} L_{id} + L_{triplet}
	\label{eq:totalLoss}
\end{equation}
where $\lambda_{l1}$, $\lambda_{adv}$ and $\lambda_{id}$  denotes the weight of L$_1$, adversarial and identity loss respectively. We set $\lambda_{l1}=0.1$, $\lambda_{adv}=0.1$ and $\lambda_{id}=10$ in our experiments.

\section{Experiments}
In these section, we report a detailed experimental analysis to validate the impact of our proposed CodeFormer++. 

\subsection{Experimental Setup}
\noindent\textbf{Implementation Details.}
We train CodeFormer++ in two stages. In the first stage, we train DAM module for alignment correction while we train TGRN module for texture injection in the second stage.
We train our model on input face image of resolution $512 \times 512 \times 3$. We employ Adam optimizer~\cite{kingma2014adam} for both the stages with batch size of 8. The initial learning rate of the optimizer is set to $5 \times 10^{-4}$. We train our model for a total of 400k iterations for stage-1 and 600k iterations for stage-2. Our method is implemented with PyTorch framework and trained using four NVIDIA Tesla V100 GPUs. 

\begin{table}[t]
	\centering
	\setlength{\tabcolsep}{0.8mm}
		\begin{tabular}{l|c|c|c|c|c|c}
			\hline
			{Methods} & PSNR↑ & SSIM↑ & NIQE↓ & LPIPS↓ & FID↓& LMD↓ \\
			\hline
			GPEN  &  21.26 & 0.565 &\underline{4.020} & 0.349 & 59.70& 7.26\\
			GFPGAN  & \textbf{25.08} & 0.677 & 4.077 & 0.365 & 42.62& 9.50 \\
			CodeFormer  &  22.18 & 0.610 & 4.520 & \textbf{0.299} & 60.62& \textbf{5.38}\\
			VQFR  & 24.14 & 0.636 & 3.693 & 0.351 & 41.28 & 9.13\\
			RF  & 24.42 & 0.640 & 4.201 & 0.365& 41.45& 8.88\\
			RF++  & 24.40 & 0.630 & 4.120 & 0.362 & \underline{38.41}& 8.52 \\
			DR2  & 23.55 & 0.595 & 4.202 & 0.434 & 50.13& 8.69 \\
			PGDiff  &22.95&0.662&4.465&0.392& 45.32& 8.71\\
			DiffBIR  &24.92 &0.675&4.060&0.477&43.82&6.18\\
			DifFace  & 23.44 & \underline{0.690} & \textbf{4.010} & 0.461 & 48.98& 6.06\\
			DAEFR  & 19.92& 0.553 & 4.477 & 0.388& 52.06& 5.63\\ 
			\hline
			\textbf{Ours}  &\underline{24.96} &\textbf{0.697}& 4.052 &\underline{0.341} &\textbf{38.13}& \underline{5.41}\\ 
			\hline
			
		\end{tabular}
		\caption{Quantitative comparisons on CelebA-Test dataset. The Best and Second Best results are highlighted in \textbf{Bold} and \underline{Underline}, respectively. Note: RF and RF++ represents RestoreFormer and RestoreFormer++ respectively.}
		\label{table1}
	\end{table}
	
	\begin{table}[t]
		\centering
		\setlength{\tabcolsep}{0.8mm}
			\begin{tabular}{l|ll|ll|ll}
				\hline
				\multicolumn{1}{c|}{Dataset} &  \multicolumn{2}{c|}{LFW-Test} & \multicolumn{2}{c|}{WebPhoto} & \multicolumn{2}{c}{WIDER-Test} \\
				{Methods} & FID↓ & NIQE↓ & FID↓ & NIQE↓ & FID↓ & NIQE↓ \\
				\hline
				GPEN  &  57.58 &3.902  &81.77 & 4.457 & 46.99&4.104\\
				GFPGAN  & 49.96 &3.882 & 87.35 & 4.144 &39.73 &3.885 \\
				CodeFormer  &  52.02 &4.482  & 78.87 & 4.550 & 39.06 & 4.164\\
				VQFR  & 50.64 & 3.589 & 75.38 & \textbf{3.607} &44.16 & \textbf{3.054}\\
				RF  & 47.75 &4.168 & 77.33 &4.587 &49.84 &3.894\\
				RF++  & 48.48 &3.960  & 74.21 & 4.204 &40.86 &3.557 \\
				DR2  & 47.93 &5.150 & 108.81 &4.782  & 47.48 &5.188\\
				PGDiff  &47.01 &4.013&82.23&4.456&39.56&4.213 \\
				DiffBIR  &\underline{46.72} & 3.972 & 81.23 & 4.412 & 38.17 & 4.182\\
				DifFace  & 46.80 &4.040  & 81.60 &4.585  & 37.52&4.240\\
				DAEFR  & 47.53& \underline{3.552} & \underline{80.13} &4.131 & \underline{36.72} & 3.655 \\
				\hline
				\textbf{Ours}  &\textbf{45.63} &\textbf{3.518}& \textbf{72.91} & \underline{3.822} & \textbf{35.21}&\underline{3.482}\\ 
				\hline
				
			\end{tabular}
			\caption{Quantitative comparisons on real-world datasets. The Best and Second Best results are highlighted in \textbf{Bold} and \underline{Underline}, respectively. Note: RF and RF++ represents RestoreFormer and RestoreFormer++ respectively.}
			\label{table2}
		\end{table}
		
		\begin{figure*}
			\centering
			\includegraphics[ width=0.9\textwidth]{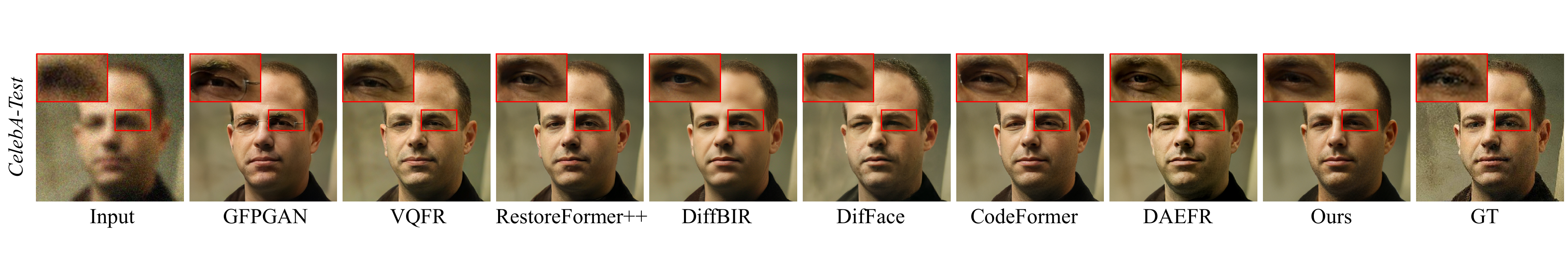}
			\vspace{-0.5cm}
			\caption{Qualitative comparisons on \textbf{CelebA-Test} dataset. \textbf{Zoom in for best view.}}
			
			\label{fig:celeb}
		\end{figure*}

		\noindent\textbf{Training Dataset.} We train our model on standard FFHQ
		dataset ~\cite{karras2019style}, consisting of 70,000 high-quality images. During training, images are resized from $1024 \times 1024$ to $512 \times 512$ resolution.
		Similar to prior arts, we generate paired dataset by synthetically corrupting clean images using the degradation pipeline ~\cite{li2020blind, wang2021towards} modeled as below :
		\begin{equation}
			I_{LQ} =\{[(I_{HQ} \otimes k_{\sigma})\downarrow _r + n_{\delta}] _{JPEG_{q}}\}\uparrow _r
			\label{eq:important}
		\end{equation}
		The high-quality image $I_{HQ}$ is first convolved with a Gaussian
		blur kernel $k_{\sigma}$, followed by a downsampling operation $\downarrow$ with
		a scale factor $r$. Subsequently, additive white Gaussian noise
		$n_{\delta}$ is added to the blurred and downsampled image. The resulting image is then  
		JPEG compressed with quality factor $q$. Finally, the degraded image is resized back to $512 \times 512$. For each training pair, we randomly sample $\sigma$, $r$, $\delta$ and $q$ from [1, 15], [1, 6], [0, 25], [30, 90], respectively. 
		
		\begin{figure*}[t]
			\centering
			\includegraphics[ width=0.9\textwidth]{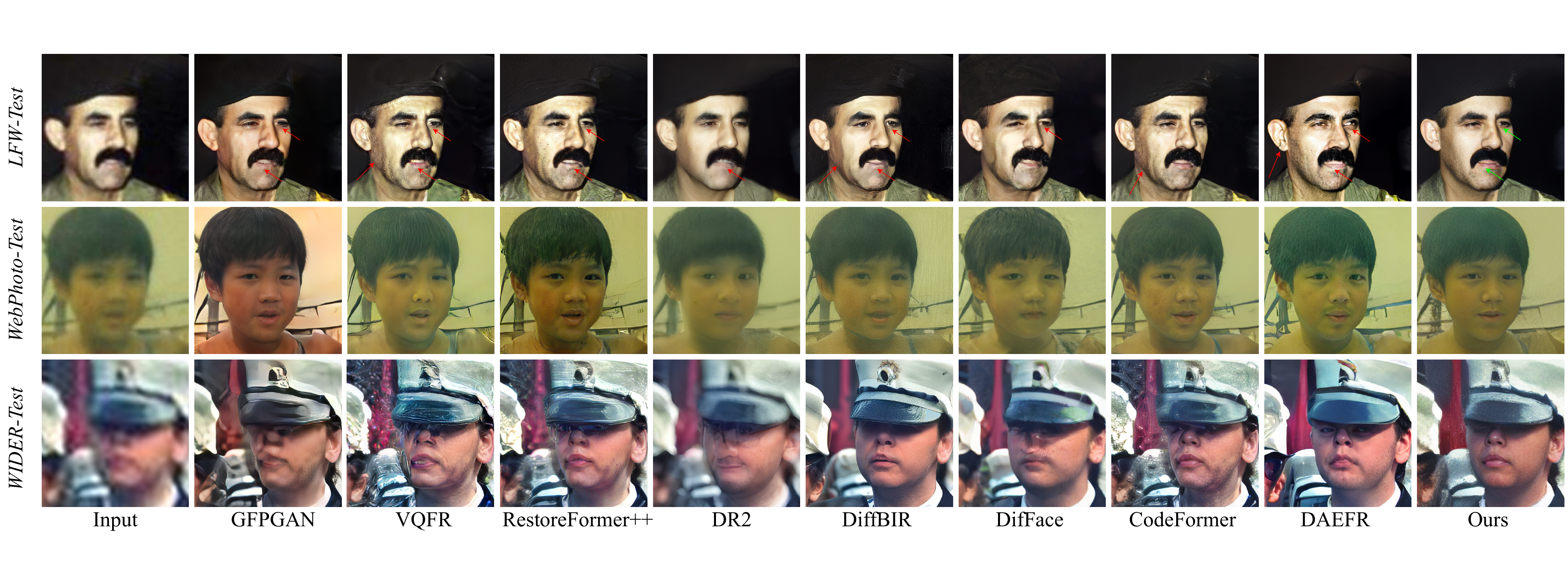}
			\vspace{-0.5cm}
			\caption{Qualitative comparisons on \textbf{LFW-Test, WebPhoto-Test and WIDER-Test} datasets. \textbf{Zoom in for best view.}}
			
			\label{fig:lfw}
		\end{figure*}

		\noindent\textbf{Testing Dataset.} We evaluate the effectiveness of the CodeFormer++ on the synthetic CelebA-Test dataset and three real-world datasets including LFW-Test, WebPhoto-Test, and WIDER-Test. CelebA-Test is a synthetic dataset with 3,000 CelebA-HQ images~\cite{karras2017progressive} and the degradation pipeline is similar to training dataset. LFW-Test~\cite{huang2008labeled} contains 1,711 real-world low-quality images. We consider the first image for each identity in the validation set of LFW dataset. WebPhoto-Test~\cite{wang2021towards}, consists of 407 low-quality faces collected from the Internet with diverse degradations. WIDER-Test consists of 970 severely degraded face images from the WIDER face dataset~\cite{yang2016wider}.
		
		\noindent\textbf{Metrics.} For quantitative evaluation, we adopt pixel-wise metrics (PSNR and SSIM) and the perceptual metric (LPIPS) for CelebA-Test where GT images are available. We also employ no-reference perceptual metrics (FID and NIQE) together with landmark distance (LMD) to effectively measure the identity distance.
		
		\subsection{Comparisons with State-of-the-art Methods}
		We compare the proposed method against several SOTA face restoration methods: GFPGAN~\cite{wang2021towards}, GPEN~\cite{yang2021gan}, RestoreFormer~\cite{wang2022restoreformer}, RestoreFormer++~\cite{wang2023restoreformer++}, DR2~\cite{wang2023dr2}, PGDiff~\cite{yang2023pgdiff}, DiffBIR~\cite{lin2024diffbir}, DifFace~\cite{yue2024difface}, CodeFormer~\cite{zhou2022towards}, VQFR~\cite{gu2022vqfr} and DAEFR~\cite{tsai2023dual}. 
		

		\noindent\textbf{Synthetic Dataset Evaluation.} 
		We present quantitative comparison on CelebA-Test dataset in Table \ref{table1}. Our proposed method, CodeFormer++, outperforms existing  methods in terms of perceptual quality metrics like FID (best score), LPIPS (second-best score), and NIQE (third-best score), indicating strong similarity between output image distribution and natural image distribution. At the same time, the proposed method exhibit competitive performance on fidelity based metric, LMD, achieving second-best score compared with other methods. It is worth noting that existing methods that perform well on perceptual metric like NIQE (GPEN, DifFace) and FID (Restorformer++), suffer from poor LMD score, indicating loss of identity information. On the contrary,  solutions such as CodeFormer, DAEFR achieve competitive LMD score but fail to restore realistic results which is quite evident from poor FID and NIQE scores.
		
		We further display in  Fig. \ref{fig:celeb} the qualitative results to support our claim. It can be clearly seen that our method is able to restore the low quality face images without deviating from the identity while producing realistic facial details. In contrast, existing methods either generate artificial spectacles (GFPGAN, DifFace, CodeFormer) or hallucinates facial features (VQFR, Restorformer++, DAEFR).
		
		\begin{figure*}
			\centering
			\includegraphics[width=1\textwidth]{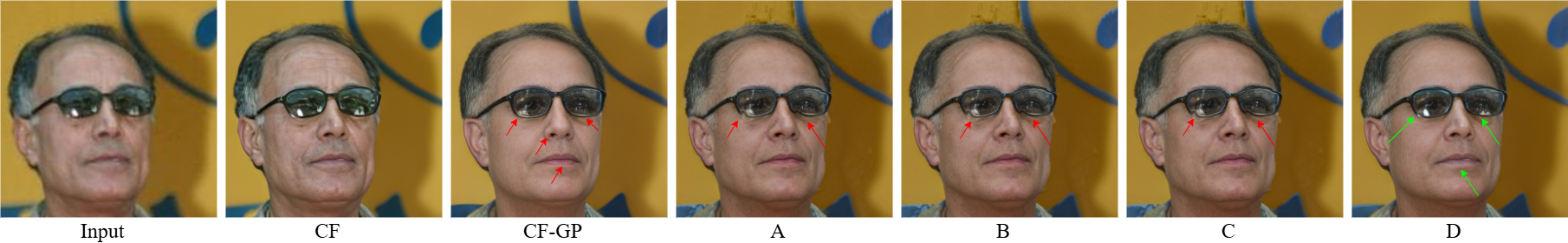}
			\caption{Ablation studies. The experimental index in accordance with the Table \ref{table3} configuration is utilized.}

			\label{fig:ablation}
		\end{figure*}

		\begin{table}[t]
			\centering
			\begin{tabular}{c|c|c|c|c|c}
				\hline
				{Metrics}&{CF-GP}&{A}&{B} &{C} &{D} \\
				\hline
				{NIQE ↓}&{4.134}&{4.136}&{4.132}&{4.112} &{4.052}  \\
				
				{LMD ↓}&{6.28}&{5.72}&{5.69}&{5.68}&{5.41} \\
				\hline
			\end{tabular}
			\caption{Ablation studies of the proposed CodeFormer++ on CelebA-Test dataset. ``A" represents deformable image alignment module. ``B" denotes TGRN trained with traditional losses. ``C" represent TGRN trained with deep metric learning with GT as a positive sample. ``D" symbolize our CodeFormer++.  }
			\label{table3}
		\end{table}

		\textbf{Real-world Datasets Evaluation.} We report in Table \ref{table2} the quantitative analysis of various methods on three different real-world datasets. It can be noticed that our CodeFormer++ achieves superior performance on all three datasets. The most encouraging finding is that our method outperforms all other methods and attain lowest FID score on all datasets. This indicates high similarity between distribution of real and generated images. In terms of NIQE metric, our method achieves the highest score on the LFW-Test dataset while attaining second-highest score on the WebPhoto-Test and WIDER-Test datasets. This vividly demonstrate the ability of our method in producing visibly pleasant results while preserving the identity. 
		
		Visual comparison in Fig. \ref{fig:lfw} further demonstrate the ability of CodeFormer++ in restoring high-quality images without altering identity. Interestingly, although VQFR obtains best NIQE score on WebPhoto and Wider-Test datasets, it significantly alters face component structures and unable to remove complex degradations. In similar way, DiffBIR~\cite{lin2024diffbir} and DAEFR~\cite{tsai2023dual} achieve second-best FID score, but they lack in recovery of facial components, leading to identity loss.

		\subsection{Ablation Studies}
		We perform several ablations to signify the importance of each module of CodeFormer++. The findings of our investigation has been presented in Table \ref{table3} and Fig. \ref{fig:ablation}.
		
		\noindent\textbf{Deformable image alignment module.} 
		The usefulness of DAM is evident with improvement in LMD score as seen in Table \ref{table3}. However, it is noticed that DAM induces artifacts on the output, especially in semantic components such as eyes, mouth and nose as observed in Fig. \ref{fig:ablation}. This is because DAM aims to establish a dense, non-linear correspondence between pair of images without guaranteeing semantically and perceptually consistent facial attributes. Hence, we cannot generate high-quality and high-fidelity facial output solely relying on DAM.      
		
		\noindent\textbf{TGRN with adversarial loss.}
		To alleviate issues associated with DAM, we train TGRN with combination L$_1$, adversarial, identity and perceptual losses. However, since artifacts are highly localized, it is difficult to discriminate between artifacts and realistic details in DAM output. Thus, traditional losses inevitably forces the network to be biased towards heavily copying facial features from DAM output, without resolving existing artifacts.           
		
		\noindent\textbf{TGRN with adversarial and triplet loss}
		To improve the discriminative power of TGRN, we integrate deep metric learning framework in the proposed work conditioned on DAM output as a negative sample and GT as a positive sample. However, we observe that issues associated with DAM output are still persistent. This is because DAM output is based on discrete codebook  which cannot model complex continuous GT  distribution precisely. This difference between discrete and continuous space makes DAM output easily distinguishable from GT in feature embedding space, making deep metric learning ineffective.     
		
		\noindent\textbf{Novel anchor positive for deep metric learning.}
		In order to effectively apply deep metric learning paradigm, it is essential to select positive and negative samples that are difficult to distinguish. In this direction, we propose to use a novel positive sample obtained by fusing facial components from CF-ID on DAM output as illustrated in Eq. \ref{eq:combination} and Fig. \ref{fig:network}. This enables the network to synergystically fuse identity and rich facial features, resulting in visually pleasing high-fidelity output which can be visualized from Fig. \ref{fig:ablation}. 
		
		\begin{table}[t]
			\centering
			\begin{tabular}{|l|c|c|}
				\hline
				{Methods} & NIQE↓ &   LMD↓ \\
				\hline
				DAEFR & 4.477&5.63\\
				DAEFR + Ours & 4.481 &5.44\\
				\hline
				RestoreFormer &4.201 &8.88\\
				RestoreFormer + Ours &4.193 &5.47\\
				\hline
				DifFace &4.010 &6.06\\
				DifFace + Ours &3.982 &5.46\\
				\hline
				
			\end{tabular}
			\caption{Extension results on CelebA-Test dataset.}
			\label{table4}
		\end{table}
		\begin{figure}[t]
			\centering
			\includegraphics[width=\linewidth]{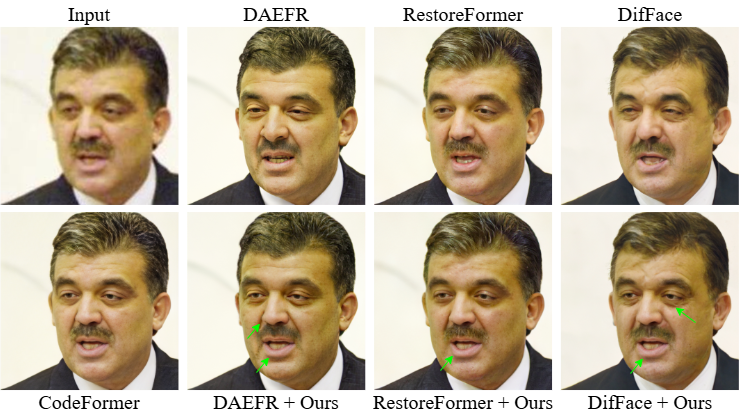}
			\caption{Qualitative comparison using DAEFR, RestoreFormer, and DifFace as a prior. \textbf{Zoom in for best view.}}

			\label{fig:extention}
		\end{figure}
		
		\subsection{Generalization}
		We demonstrate the generalizability of our framework by extending it to other generative prior and transformer based methods that heavily suffer from fidelity. To do this, we replace CF-GP output with DAEFR (~\citeauthor{tsai2023dual}), RestoreFormer (~\citeauthor{wang2022restoreformer}), and DifFace (~\citeauthor{yue2024difface}) outputs. From visual results showcased in Fig. \ref{fig:extention}, the remarkable improvement in identity  without compromising on textural details, across all methods can be clearly witnessed. This generalizability can also be vividly  seen from Table \ref{table4} with significant reduction in LMD score with negligible change in NIQE scores.

		\section{Conclusion}
		We propose CodeFormer++, a novel framework for BFR that effectively balances identity preservation with realistic texture reconstruction. To this end, the DAM first aligns
		the generative prior and with identity-preserving restored image. These aligned representations are then adaptively fused by TGRN to generate visually plausible and identity-consistent face images. This process is reinforced by deep metric learning to ensure identity fidelity. Extensive experiments on both synthetic and real-world datasets demonstrate the superiority of our approach, establishing a new benchmark in BFR.

		
		

		\bibliography{main}
\newpage
        \section{CodeFormer++: Supplementary Material}
        
\begin{figure}[b]\setlength{\hfuzz}{1.1\columnwidth}
	\hspace{-250pt}
	\begin{minipage}{\textwidth}
		
		\begin{center}
			\includegraphics[width=0.9\linewidth]{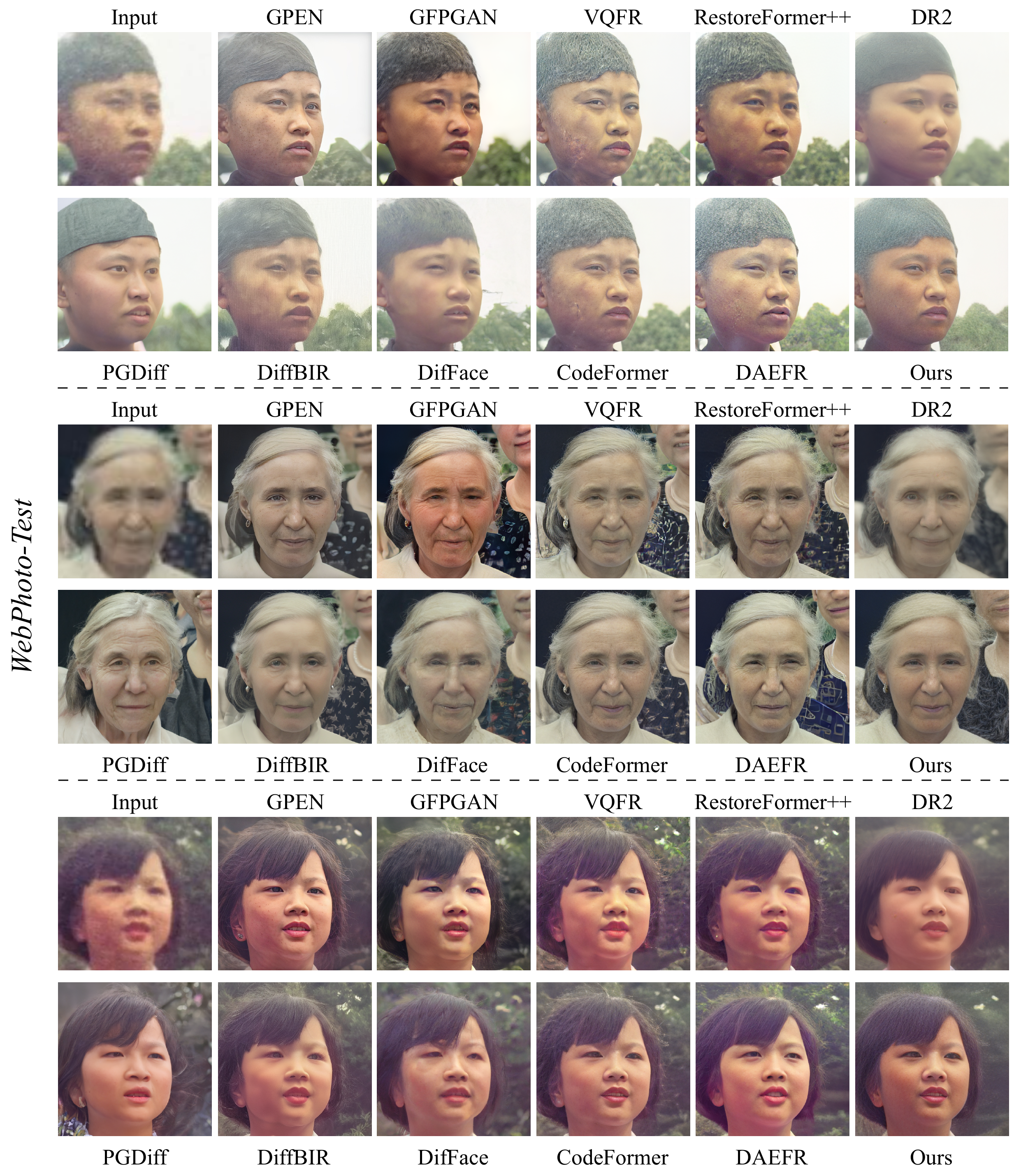}
		\end{center}
		\caption{Qualitative results on \textbf{WebPhoto-Test} dataset. Our method is able to effectively reconstruct identity consistent high-texture faces when compared to SOTA, across various levels of degradation.
		}
		\label{fig:sup_webphtot}
	\end{minipage}
\end{figure}


\begin{figure*}[t]
	\centering
	\includegraphics[width=\textwidth]{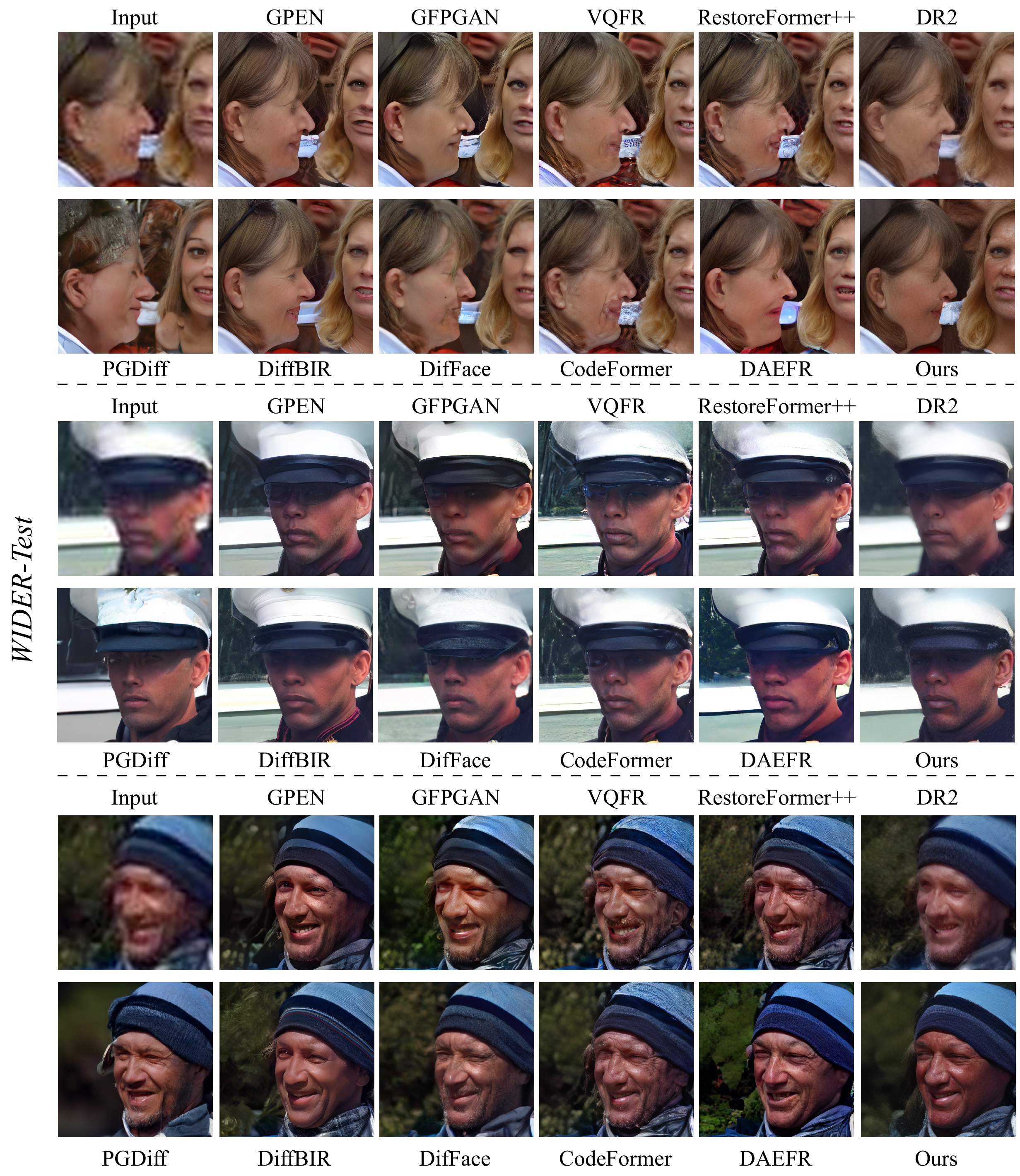}
	\caption{Qualitative results on \textbf{WIDER-Test} dataset. Our method is able to effectively reconstruct identity consistent high-texture faces when compared to SOTA, across various levels of degradation.}

	\label{fig:supp_wider1}
\end{figure*}
\begin{figure*}[t]
	\centering
	\includegraphics[width=\textwidth]{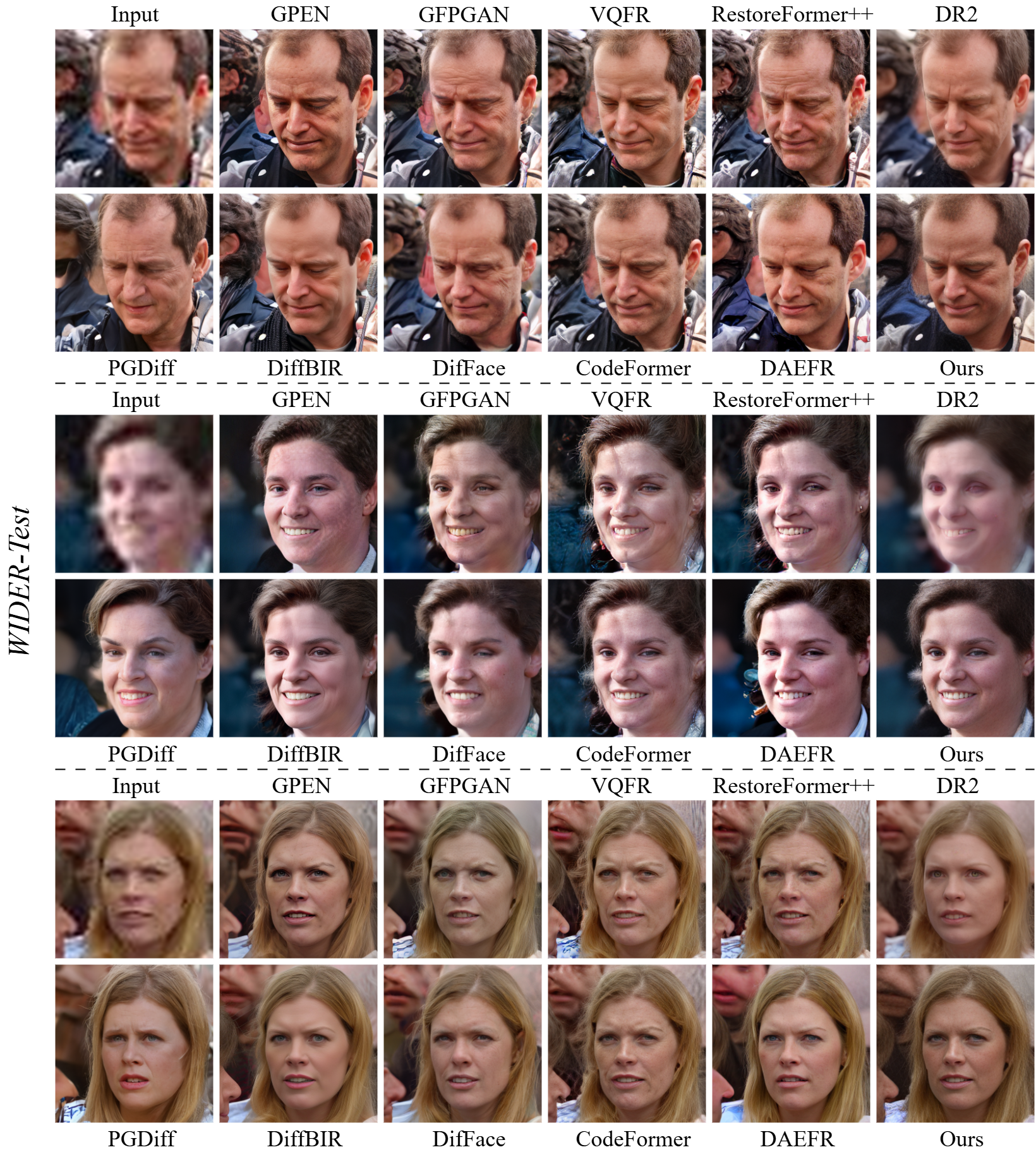}
	\caption{Qualitative results on \textbf{WIDER-Test} dataset. Our method is able to effectively reconstruct identity consistent high-texture faces when compared to SOTA, across various levels of degradation.}

	\label{fig:supp_wider2}
\end{figure*}
		
	\end{document}